\documentclass[lettersize,journal,review]{IEEEtran}
\usepackage{amsmath,amsfonts}
\usepackage{algorithmic}

\usepackage{array}
    \newcolumntype{P}[1]{>{\centering\arraybackslash}p{#1}}
    \newcolumntype{M}[1]{>{\centering\arraybackslash}m{#1}}
    
\usepackage[caption=false,font=normalsize,labelfont=sf,textfont=sf]{subfig}
\usepackage{textcomp}
\usepackage{stfloats}
\usepackage{url}
\usepackage{verbatim}
\usepackage{graphicx}
\usepackage{cite}
\usepackage{enumitem}
\usepackage{multirow}
\usepackage{longtable}
\usepackage{float}
\usepackage{xcolor}

\usepackage{pdflscape}
\usepackage{afterpage}

\def\BibTeX{{\rm B\kern-.05em{\sc i\kern-.025em b}\kern-.08em
    T\kern-.1667em\lower.7ex\hbox{E}\kern-.125emX}}
\usepackage{balance}

\newtheorem{definition}{Definition}

\begin{document}
\title{Explainability in reinforcement learning: perspective and position}

\author{Agneza Krajna
\IEEEmembership{Member, IEEE}, Mario Brcic
\IEEEmembership{Member, IEEE}, Tomislav Lipic
\IEEEmembership{Member, IEEE}, Juraj Doncevic
\IEEEmembership{Member, IEEE}
\\
\textbf{\textcolor{red}{ This work has been submitted to the IEEE for possible publication. Copyright may be transferred without notice, after which this version may no longer be accessible.}}
\thanks{A. Krajna, M. Brcic, and J. Doncevic are with University of Zagreb Faculty of Electrical Engineering and Computing, Zagreb,
Croatia.}
\thanks{T. Lipic is with Institute Ruder Boskovic, Zagreb, Croatia.}
\thanks{A. Krajna is corresponding author at: agneza.krajna@fer.hr}
}

\markboth{Journal of XXXX Files,~Vol.~18, No.~9, September~2020}%
{}

\maketitle

\begin{abstract}
Artificial intelligence (AI) has been embedded into many aspects of people’s daily lives and it has become normal for people to have AI make decisions for them. From helping users to find their favorite items to purchase, recommending movies and friends on Facebook, to life-essential decisions. Reinforcement learning (RL) models increase the space of solvable problems with respect to other machine learning paradigms. Some of the most interesting applications are in situations with non-differentiable expected reward function, operating in unknown or underdefined environment, as well as for algorithmic discovery that surpasses performance of any teacher, whereby agent learns from experimental experience through simple feedback. The range of applications and their social impact is vast, just to name a few: genomics, game-playing (chess, Go, etc.), general optimization, financial investment, governmental policies, self-driving cars, recommendation systems, etc. It is therefore essential to improve the trust and transparency of RL-based systems through explanations. 
Most articles dealing with explainability in artificial intelligence provide methods that concern supervised learning and there are very few articles dealing with this in the area of RL.
The reasons for this are the credit assignment problem, delayed rewards, and the inability to assume that data is independently and identically distributed (i.i.d.).
This position paper attempts to give a systematic overview of existing methods in the explainable RL area and propose a novel unified taxonomy, building and expanding on the existing ones. The position section describes pragmatic aspects of how explainability can be observed. The gap between the parties receiving and generating the explanation is especially emphasized. To reduce the gap and achieve honesty and truthfulness of explanations, we set up three pillars: proactivity, risk attitudes, and epistemological constraints. To this end, we illustrate our proposal on simple variants of the shortest path problem.    

\end{abstract}


\section{Introduction}
\begin{figure*}[ht]
\begin{minipage}[b]{0.45\linewidth}
 \centering
    \includegraphics[width=3.5in]{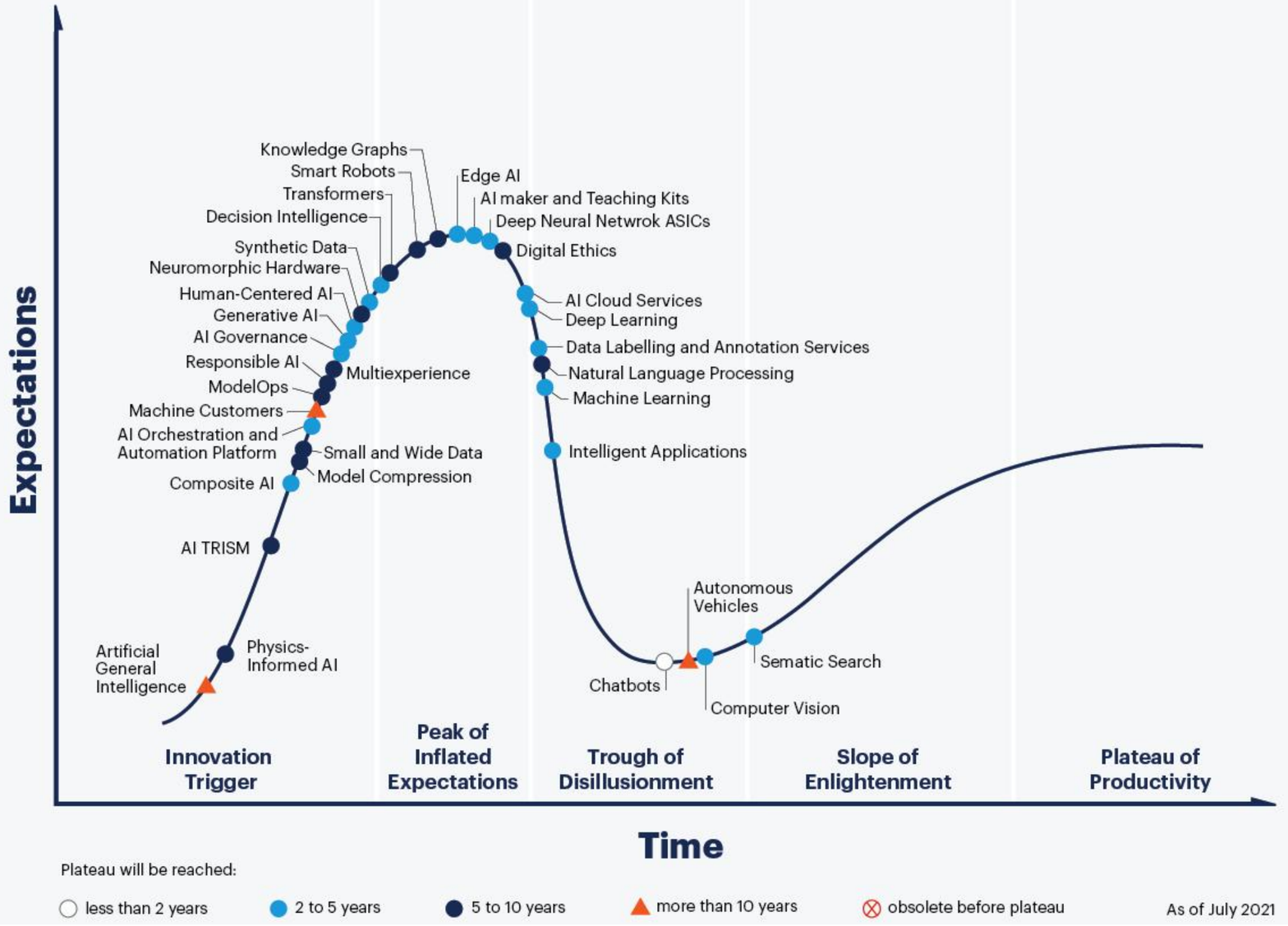}
    \caption{ Hype Cycle for Artificial Intelligence 2021 \cite{noauthor_gartner_nodate}}
    \label{fig:hype_cycle_ai}
\end{minipage}
\hspace{0.5cm}
\begin{minipage}[b]{0.45\linewidth}
    \centering
    \includegraphics[width=3.5in]{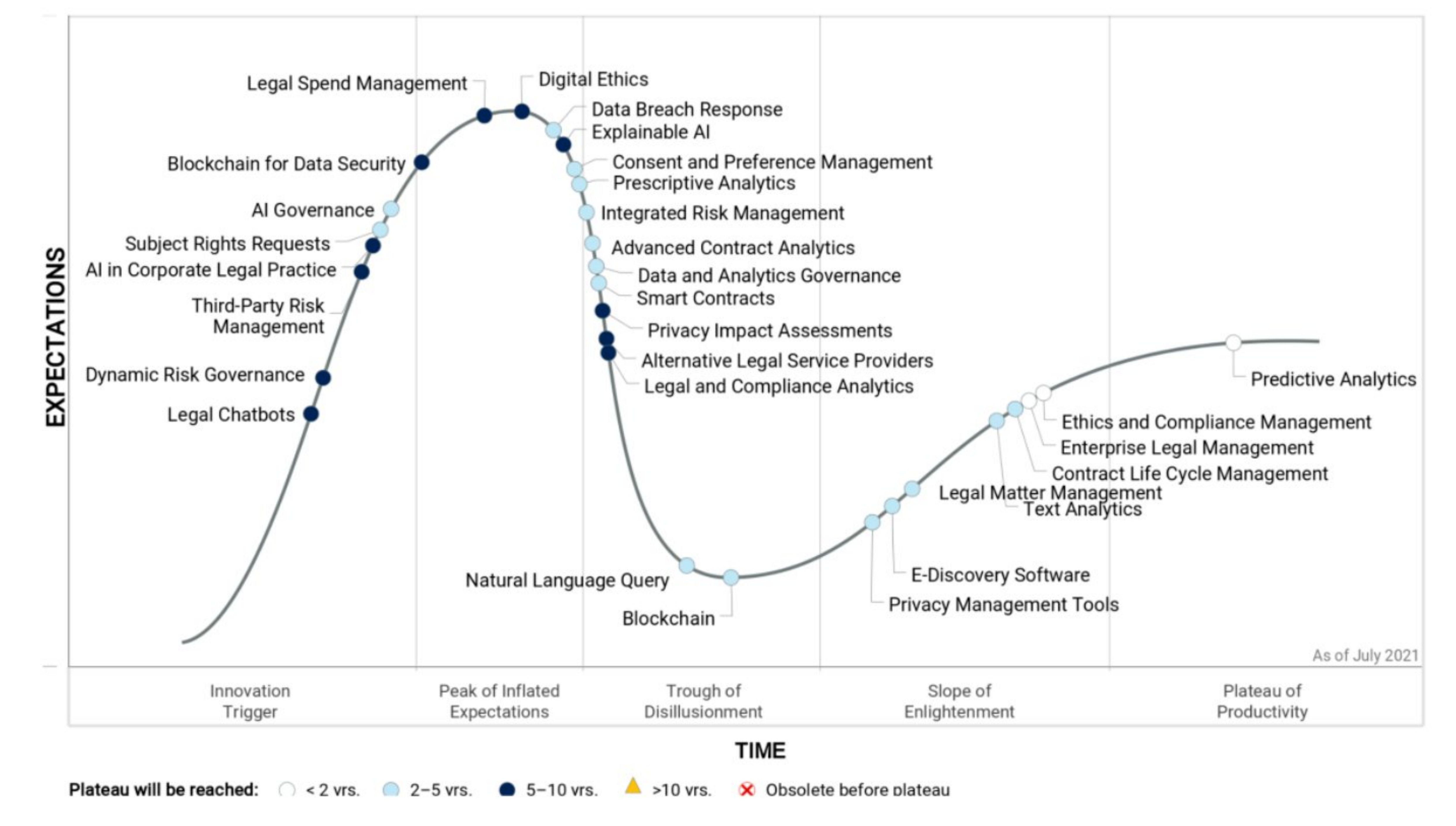}
    \caption{\centering Gartner Hype Cycle for Legal and Compliance Technologies 2021\cite{noauthor_gartner_nodate-1}}
    \label{fig:hype_cycle_legal}
\end{minipage}
\end{figure*}

In the last decade, as computing power and disk storage capacity increased, there has been a significant boost in the capabilities of Artificial Intelligence (AI). Due to advances in Machine Learning (ML) and autonomous decision making, systems based on machine learning have achieved superhuman performance in performing various tasks such as image processing, speech recognition, strategic planning. However, even with the remarkable performance of intelligent systems, their lack of transparency and comprehensibility remains a key obstacle for their adaptation in more broader human-centric safety-critical application scenarios. Since AI systems already have a significant impact in everyday life, the issue and development of the field of eXplainable Artificial Intelligence (XAI) is becoming increasingly important. Explanations are used for \cite{dosilovic_explainable_2018, adadi_peeking_2018}:
\begin{enumerate}
\item {Justification - to  answer a question of why a decision was made in order to increase trust in the model.}
\item {Monitoring and debugging – understanding system behavior provides the ability to find faults and their causes and solve them.}
\item{Improving – understanding how the system works provides the ability to make it smarter and enhanced.} 
\item{Discovering – Some AI systems outperform humans in solving tasks. A notable example is chess. If the system could give an explanation, one could discover and learn new things.}
\end{enumerate}
Explanations can be utilized in design and development of algorithms. For example, explanations have been used to find the conceptual flaws in the assumptions of existing optimization approaches to design a better approximate dynamic programming algorithm \cite{brcic_tracking_2018, brcic_planning_2019}.
Furthermore, in some application domains, in some fields, failure is not an option. Momentary deficiency in computer vision algorithms can easily lead to fatality in autonomous vehicles. In the medical domain, human lives are also at stake. Detection of  disease at its early phase is often critical to the recovery of patients or to prevent the disease from advancing to more severe stages \cite{tjoa_survey_2021}. These are reasons why it is essential to improve the trust and transparency of the AI system to ensure that decisions, especially those that are important for life, are made correctly. There is a special branch of AI called AI safety that deals with the question of how to make AI safe for humans. This field \cite{juric_ai_2020}  is divided into three subfields: technical AI safety, AI ethic and AI policy. The field of explainability itself is located in the subcategory of technical security called assurance. 
According to Gartner’s Hype Cycle for Artificial Intelligence 2021 (Figure \ref{fig:hype_cycle_ai}), the area of the field of Responsible Artificial Intelligence \cite{barredo_arrieta_explainable_2020} is currently in the Innovation of Trigger phase. This field is one of the AI mega trends that includes explainable AI, risk management and AI ethics for increased trust and transparency of AI. This field is projected to reach its Plateau of Productivity in a period of 5 to 10 years. 

This interest is partly due to the European Union’s General Data Protection Regulation (GDPR) that went into effect on May 25, 2018.   According to this regulation, the subject has the right not to be subject to a decision based solely on automatic processing and, if he has consented to the said processing, to obtain an explanation for the decision and the right to challenge it \cite{goodman_european_2017}. 
On April 21, 2021, the European Commission proposed an AI regulation, The Artificial Intelligence Act  \cite{noauthor_proposal_2021}. This act addresses the risks stemming from the various uses of AI systems and creates new regulatory obligations for AI tools used in various services like education, finance and law enforcement, medical devices, toys, and so on. It is being implied that EU regulations on artificial intelligence will work in line with the GDPR and will not attempt to change data privacy regulations. Therefore, it is not only desirable, but also an often requirement, that systems can explain their decisions and that they are in no way biased.

Consequently, the field of eXplainable AI (XAI) has emerged as one of the most  popular and fruitful AI research fields, which is demonstrated by numerous reviews and surveys being published in the recent years \cite{guidotti_survey_2018,  arya_one_2019, ras_explainable_2021, burkart_survey_2021, gerlings_reviewing_2021,  guidotti_principles_2021, barredo_arrieta_explainable_2020,  dosilovic_explainable_2018, tjoa_survey_2021}. However, most of these reviews and surveys on XAI were mostly exclusively focused on concepts and taxonomies of techniques suited for the supervised learning paradigm, neglecting specific peculiarities needed for easier and reliable translation of these efforts to the reinforcement learning paradigm and applications. Furthermore,  there are numerous practical examples that each particular ML problem setting (type of data and type of learning model) demands for its own custom-tailored explanation concepts and tools in order to optimally address explainability challenges \cite{chen_seven_2021, neely_order_2021}.  

Specifically, ML algorithms are commonly divided into three main categories based on their purpose and feedback used during the learning phase: supervised learning, unsupervised learning, and reinforcement learning.  Supervised learning involves learning from a training set of labeled examples provided by a knowledgeable external supervisor, while unsupervised learning is typically about finding latent structures hidden in collections of unlabeled data \cite{sutton_reinforcement_1998}. On the other hand, in reinforcement learning the agent is located in an environment that represents everything outside the agent. The agent interacts with the environment by performing various actions sequentially. An agent without knowing upfront which actions to take in a certain situation has to uncover the policy for taking actions by experimentation. Some action may affect not only the immediate reward but also the next situation and all subsequent rewards. The interaction happens at each sequence of discrete time steps and as a consequence of its action, one step later,  the agent receives a numerical reward \cite{sutton_reinforcement_1998}. This breaks the common i.i.d. assumption in machine learning. 

Although the popularity of XAI is at its peak, most research is oriented on interpretability and explainability in supervised learning algorithms, while research on explainability in Reinforcement learning (XRL) has just started to gain its momentum as a distinct XAI research domain. The conceptual  nature and working principles of RL algorithms make XRL inherently distinct and more challenging  than XAI on supervised learning \cite{dazeley_explainable_2021}. Most current techniques are based on different concepts and are focused on different aspects of explanations. Hence, this all raises the need of a systematic overview covering methods and concepts  of RL explanation techniques, and this paper tries to fill this gap.

\subsection{Existing XRL overviews and the scope of this paper}
There are only a sparse number of existing works that attempt to consolidate current emerging trends of specialized interpretable ML and XAI techniques suited for reinforcement learning paradigm \cite{alharin_reinforcement_2020,dazeley_explainable_2021,glanois_survey_2021,heuillet_explainability_2021,puiutta_explainable_2020}, and our paper adds to this body of work. 

In 2020, Puiutta and Veith in \cite{puiutta_explainable_2020}, first surveyed partly limited number of existing  XRL methods, noticing that most of XRL methods, similarly to XAI body of work, often disregard the fact that explainability should incorporate aspect of the human involvement \cite{roscher_explainable_2020} and that explanations should be human-centric and tailored for a user with specific level of knowledge and expertise \cite{barredo_arrieta_explainable_2020,arya_one_2019}. Although it is the first published survey on explainability in RL, their classification and overview of selected XRL models were provided only based on XAI taxonomy and categories. Similarly, other existing subsequent works on surveying XRL approaches also still relied on following established concepts, definitions and terminology derived from XAI and interpretable ML \cite{alharin_reinforcement_2020,glanois_survey_2021,heuillet_explainability_2021}. Survey \cite{glanois_survey_2021} gave overview of various XRL approaches with the main focus on using intrinsically interpretable models (by covering interpretable inputs,  approaches for learning interpretable state transition and reward model, and for learning interpretable preference model and policies) and purposely less covering post-hoc interpretability. In addition, Heuillet et al. \cite{heuillet_explainability_2021} provided a more comprehensive review of the most prominent XRL works, showing how XAI techniques are used to enable explainability of RL (and deep RL) models with  taking into account  the target audience for explanation. Therefore, in these works the overview of XRL research was still provided only based on definitions from XAI and interpretable ML field, interpretability aspect followed in \cite{glanois_survey_2021},  and full explainability aspect that considers target human audiences in \cite{heuillet_explainability_2021}.

On the other hand, Dazeley and coauthors \cite{dazeley_explainable_2021}, were the first to emphasize XRL as a distinct branch of XAI  that (i) has potential to serve as foundation of broad XAI and that (ii) goes well beyond simply interpreting decisions with usual supervised learning approaches to explanation from the area of interpretable ML. While reviewing existing approaches explaining RL agent behavior as the state of the art XRL research, the authors introduced and used  a conceptual framework for XRL based on the Causal Explanation Network (CEN) model for human lay causal explanations. The CEN model incorporates seven relevant categories for causal reasoning about agents behavior, being readily applicable to  the future developments that are specific to XRL, or even Multiagent Reinforcement Learning (MARL) domains. 

We build and extend on these starting works by systematically classifying the current state of the art of XRL research based on our newly proposed unified taxonomy that pragmatically accommodates timely and foreseeable future trends of XRL research (including explainability in MARL), while being also aligned with previous efforts in XAI research. As opposed to all current works summarizing XRL research, our work also includes and explores beyond  single-agent XRL methods, and our taxonomy can easily accommodate state of the art XRL research trends of developing unified explanation approaches for understanding agent's policy, pinpointing good policies and performing policy forensics that are based on identifying the critical time steps contributing to the final reward \cite{guo2021edge}.

\subsection{Paper contributions} 

This position paper gives a systematic overview of existing methods in the explainable RL area based on our newly introduced practical taxonomy. Specifically, the distinct cont
ributions of our paper can be summarized as follows:
\begin{itemize}
    \item New taxonomy: We propose a novel unified taxonomy of existing and potentially newly emerging explanation techniques for RL (including deep RL and causal RL) 
    \item Systematic  review: We give categorization of explainability methods for RL  depending on the number of time steps and the actions in them that affect the current decision. Each of the methods is further described by belonging to one of the following categories: type of environment in which the agent operates (deterministic/stochastic), type of policy followed by the agent (deterministic/stochastic), and the number of agents operating (single-agent / multi-agent).We systematically outline details of each explanation methods, pinpointing their advantages and disadvantages 
    \item Proposal for explainability in RL (XRL): We pose proposal  for truthful explanations based on  three pillars: explanations should be proactive, explanation must take into account risk attitudes and decision-maker’s computational process must be taken into account in the form of epistemological constraints
    \item Illustrated XRL proposal: We illustrate our proposal on simple variants of the shortest path problem through the variation of our proposed axes: type of environment, agent cardinality and policy type.
    \item Future XRL research outlook: we discuss limitations, pitfalls, and recommend development of practical RL model explainability factsheets 
\end{itemize}

\subsection{Paper structure}
Basic concepts and detailed  definitions related to explainability and  reinforcement learning paradigm are given in section \ref{sec:defs}. In section \ref{sec:tax}, we present  novel  taxonomy based on 6 aspects of explainability and systematically outline current works on explainability techniques for reinforcement learning. The section \ref{sec:prop} describes pragmatic aspects of how explainability can be observed. The gap between the parties receiving and generating the explanation is especially emphasized. To reduce the gap and achieve honesty and truthfulness of explanations, we set up three pillars: proactivity, risk attitudes and epistemological constraints.
Finally, we conclude in section \ref{sec:concl}.

\section{Basic terms and definitions}
\label{sec:defs}
\subsection{Definition of explanation}
In order to achieve cooperation between an algorithm and humans, trust is necessary, and it can only be achieved by explainability \cite{dosilovic_explainable_2018}. 
Das and Rad \cite{das_opportunities_2020} define  Explainable Artificial Intelligence (XAI) as a field of Artificial Intelligence that promotes a range of tools, techniques, and algorithms that can generate explanations that are interpretable, intuitive, and understandable for humans. According to IBM, XAI is a set of processes and methods that allow human users to trust the machine learning outputs \cite{noauthor_explainable_2021}.
Halpern et al. \cite{halpern_causes_2005} gave a different definition of explanation. They said that an explanation is a fact that is not known for the certain but, if found to be true, would constitute an actual cause of the explanandum (the fact to be explained). They take the notion of explanation to be relative to the agent’s epistemic state. This leads to one of the main problems in explainability, and that is the dependence of explanation on the domain of knowledge that the agent possesses. What is an explanation for one agent may not be an explanation for another agent \cite{halpern_causes_2005}.  For example, a person lives near a waste incinerator and they have lung problems. For someone who knows where the man lives, an explanation will be: “Floating particles, which are released by burning, can reach deep into the lungs by inhalation and can bind carcinogens to themselves”.  But for someone who already knows how living near waste incinerators dangerous is, the explanation for the person’s condition will be: “He lives near a waste incinerator”.  Similar analogy could be given from the domain expert-non-expert perspective, where, for instance, medical professionals would appreciate explanations based on specific concepts from the medical domain or their professional specializations, while patients not educated in the medical domain would probably prefer general explanations in layman's words and general concepts.

In \cite{montavon_methods_2018} an explanation is defined as a collection of features of the interpretable domain, that have contributed for a given example to produce a decision.  For example, an explanation can be a heatmap highlighting pixels of the input image which most influenced the making of a particular decision.

Explainability and interpretability are used interchangeably in many papers \cite{tjoa_survey_2021}, while in others clear or delicate differences are emphasized \cite{barredo_arrieta_explainable_2020,roscher_explainable_2020,rudin_stop_2019,yuan_explainability_2021} For instance, Roscher et al. \cite{roscher_explainable_2020} explicitly differentiate between transparency, interpretability and explainability concepts, defining that {\bf transparency} considers the ML approach (i.e. working properties and principles of learning algorithm), while {\bf interpretability} considers the ML model together with data, and finally  in addition to the model and the data, {\bf explainability} needs to consider human involvement aspect (i.e. providing human-understandable explanations for model prediction according to the level of human specific expertise). On the other hand Rudin \cite{rudin_stop_2019}  relies on more pragmatic (but somewhat restrictive) definitions,  considering interpretability as the ability of intrinsically interpretable models and explainability as an ability to explain models by using post-hoc interpretability techniques. These definitions are also used further in recent XAI surveys by \cite{yuan_explainability_2021} and \cite{glanois_survey_2021}. Finally,  in \cite{barredo_arrieta_explainable_2020}, Aerrieta et al.clarify the distinction and similarities among majority of various concepts often used in XAI communities differencing between understandability (or equivalently, intelligibility), comprehensibility, interpretability, explainability and transparency, as well as its levels simulatability, decomposability and algorithmic transparency.

For the purpose of this paper, explainability means a technical understanding of the connection between the inputs and outputs of a particular AI model. That means that a person is able to generate an output and understand how it originated from the input features. Interpretability is being able to literally explain what is happening behind the curtain. In \cite{montavon_methods_2018} interpretation is the mapping of an abstract concept into a domain humans can make sense of. So, in the ML system context, interpretability is the ability to explain or present results in understandable terms to humans \cite{dosilovic_explainable_2018}. In \cite{freitas_comprehensible_2014} comprehensibility is used as a synonym for interpretability.

\begin{figure}
    \centering
    \includegraphics[width=3.5in]{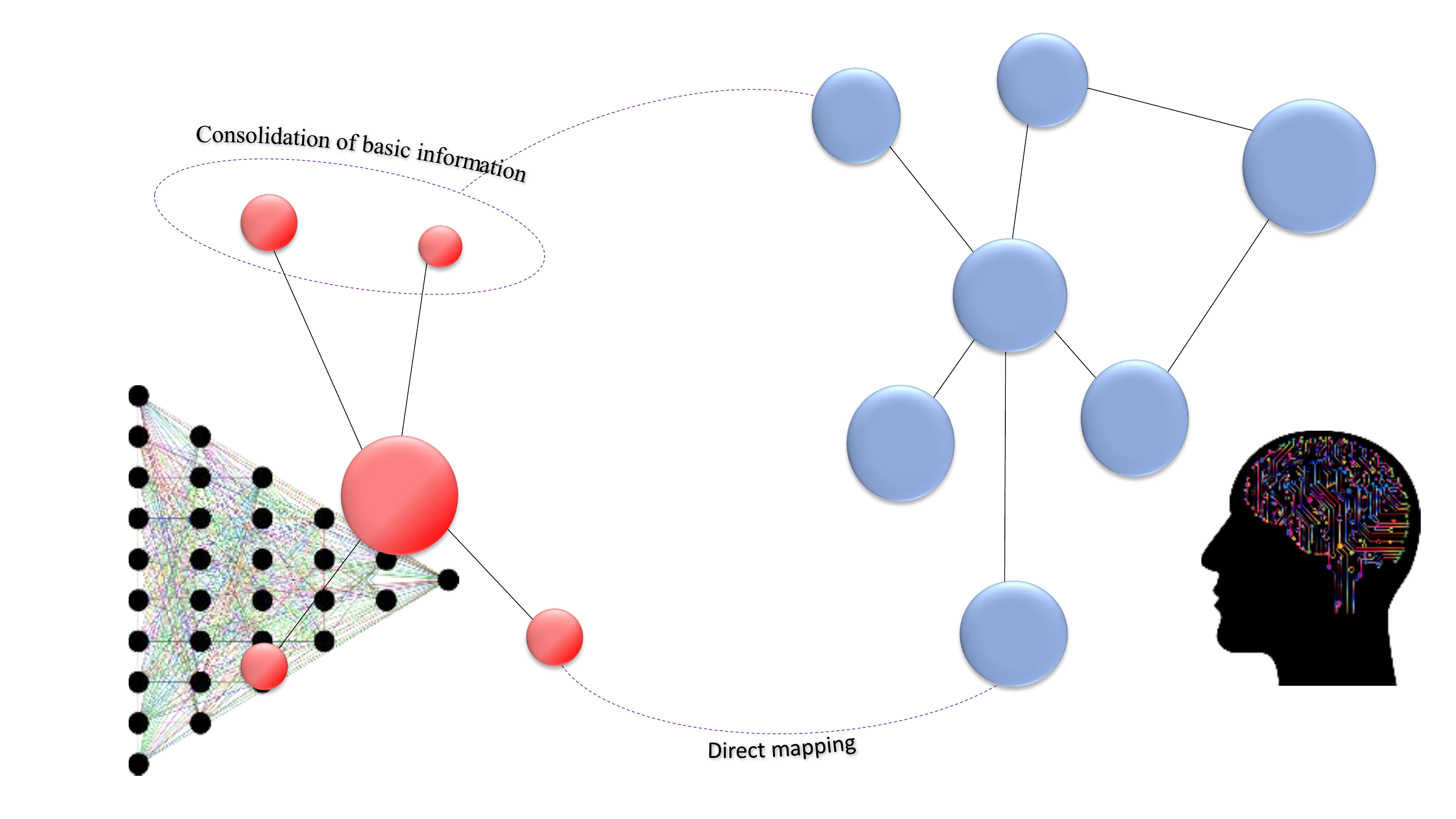}
    \caption{Conceptual presentation of explanations}
    \label{fig:concept_explain}
\end{figure}

Figure \ref{fig:concept_explain} conceptually shows the explanation mechanism. On the right, the graph of knowledge is shown in blue as a model of human knowledge. The graph consists of nodes that represent various facts stored in the human brain that are interconnected depending on their relationship. The model of knowledge generated by the computer model (e.g. neural network model) is shown on the left. For the result of a machine model to be presented to the human mind, it is necessary to translate numbers, vectors, pixels, and the like into a concept that man can understand and interpret. The manner and form of interpretation can of course also depend on the specialized domain to which the problem belongs (e.g. medical domain, economic domain, etc.).

The dashed lines in the figure represent the procedures by which the machine model of knowledge will approach the human model of knowledge by connecting external concepts with the human’s internal ones. Sometimes this connection will be realizable directly, and sometimes additional basic information will have to be connected in order to obtain a larger, more understandable concept closer to a human’s understanding (e.g. connecting more pixels in a specific segment of the image). Connecting one concept with another will allow assimilation and understanding of the computer model. The procedures by which this is achieved can be concrete images, textual descriptions, diagrams, a more understandable graphical representation of a particular decision such as a decision tree, some concrete example close to a human, and the like. 

\subsection{Reinforcement learning}

Reinforcement learning is a type of learning through trial-and-error interactions with a dynamic environment which represents the outside world programmatically. The RL framework is comprised of agents, environments, states, and actions. In the beginning the agent is in a state ($s \in S$) and it interacts with the environment. In some states the agent can choose one of several actions ($a \in A$), and as a result there is a transition to another state and a scalar reward ($r \in R$). In the beginning, the agent may even know nothing about the environment, but based on the actions and received rewards, it can find good behavioral plans.

\begin{figure}
    \centering
    \includegraphics[width=3.5in]{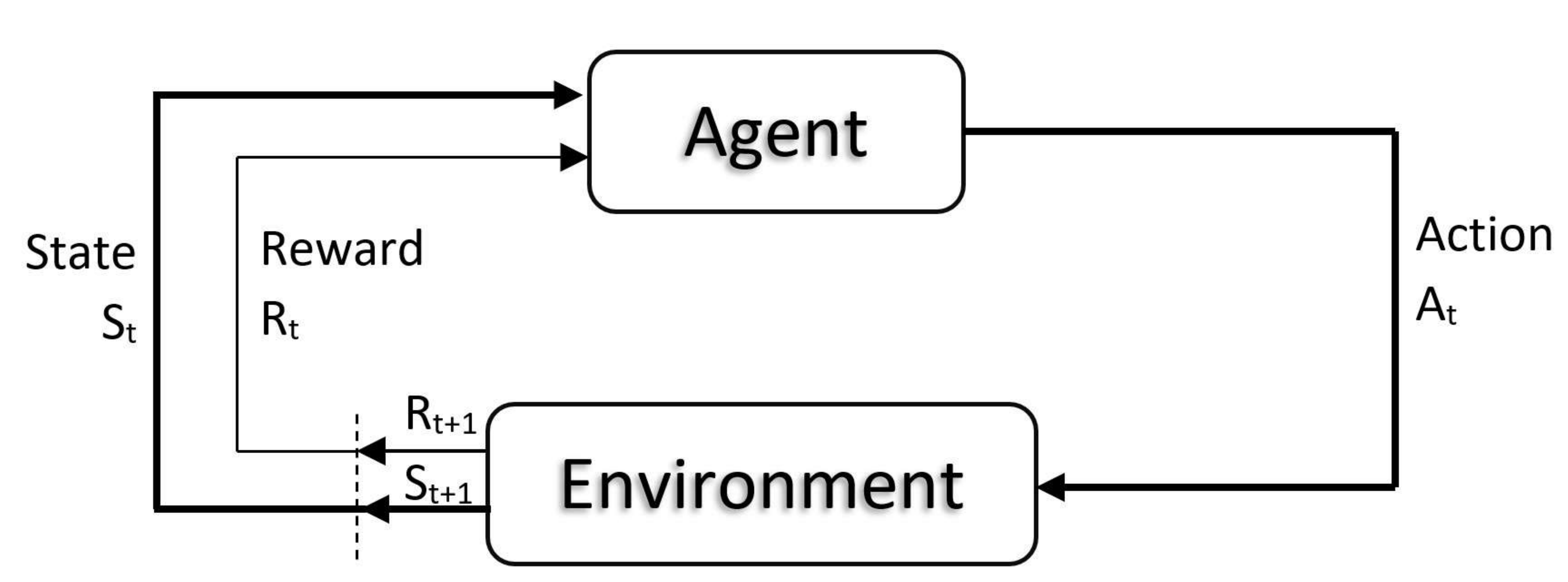}
    \caption{Typical agent - environment cycle}
    \label{fig:agent_envir}
\end{figure}

The $\pi(S)$ policy provides guidance on which action is optimal to take in a particular state in order to maximize the overall reward. Also, each state $s\in S$ is evaluated with a value function $V_\pi(s)$ that shows the amount of total reward the agent can expect if it comes to a certain state s and continues to follow the current policy $\pi(S)$. $\pi(S)$ and $V(S)$ are what we are trying to learn by RL, and there are different approaches depending on how they approach those two objects of interest. Another important function is the action-value function $Q_\pi(s,a)$ that shows, following the policy $\pi$, how good a certain action $a$ is in the state $s$.

A lot of RL problems with discrete actions are formally defined as Markov decision processes (MDP) that are defined with a tuple $(S,A,P_a,R_a)$ where $S$ stands for set of the states, $A$ for set of the actions, $P$ is the probability of transition from one state to another with action a at time $t+1$ and $R$ is an immediate reward after transition from one state ($s$)to another ($s'$) due to action a. The Bellman Optimality Equations are used to find optimal policy and thus solve MDP process by finding the optimal value $V^*$ (\ref{eq:v_value}) and/or $Q^*$s-function (\ref{eq:q_value}). The Bellman Optimality Equations for infinite horizon problems (assuming maximization problem) take into account that the utility of a condition depends on the immediate reward and also on the future reward discounted by factor $\gamma$: 

\begin{equation}
    \label{eq:v_value}
    V^*(s) = max_a \mathbb{E}_s (R(s,a,s') + \gamma V^*(s'))
\end{equation}
\begin{equation}
    \label{eq:q_value}
    Q^*(s,a) = \mathbb{E}\{R(s,a,s') + \gamma max_{a'} Q^*(s',a')\}
\end{equation}

In value-based RL approaches, the policy $\pi(s)$ can be easily indirectly generated on-the-fly from $Q^*$-function:
\begin{equation}
    \pi^*(s) = argmax_a Q^*(s,a)
\end{equation}

Value-based Deep Reinforcement Learning (DRL) uses a Deep neural network (DNN) to estimate Q-values or value function. Policy-search DRL uses DNN to directly search for a high-performing policy function $\pi(s)$.  The goal of Explainable RL is to explain how these policies and their decisions are generated in a way that is understandable to humans.

Explainability in model-free reinforcement learning is scarcely explored. The problem with RL is that all the possible combinations of states and actions are unknown, and it is difficult to cover all cases. Another problem is that neural networks, used to learn policies, are incomprehensible themselves. They are used as input-output black boxes. Recently, there has been many research that proposes environments and attempts to build tools that allow for DNN explanations and interpretations \cite{sequeira_interestingness_2020, shu_hierarchical_2017, spinner_explainer_2020, wang_reinforcement_2018}.

\section{Taxonomy and literature review}
\label{sec:tax}
This paper offers a different categorization of methods depending on the number of time steps and the actions in them that affect the current decision.
In survey articles dealing with explainability in AI, the main categorization of methods is according to the following two axes:
\begin{enumerate}[label=\textbf{A\arabic*}, leftmargin=*]
     \item \label{enum:A1}Scope of explanation - global/local methods – depending on the scope of the explanation,
    \item \label{enum:A2}Timing of explanation - intrinsic/post hoc interpretability methods – depending on the time when the explanation is generated.
\end{enumerate}

According to axis \ref{enum:A1}, the methods are divided according to the scope of explanation. Local models offer explanations for a specific decision/prediction or feature and help answer the question “Why did the model make this decision in this example” \cite{puiutta_explainable_2020}.  In contrast, the global method offers an explanation for general model behavior \cite{puiutta_explainable_2020}, across all predictions. It considers which features contribute and influence the model’s decisions and predictions the most.

According to axis \ref{enum:A2}, the methods are divided according to the timing of explanation.Intrinsically interpretable ML models are constructed to be inherently interpretable or self-explanatory during the training time by restricting the complexity of the model while in post-hoc methods explainability is achieved after training by creating a simplified surrogate simpler model \cite{puiutta_explainable_2020}. Intrinsic models are considered interpretable due to their simple structure like decision tree, and post hoc refers to the application of interpretation methods after model training like permutation feature. Post hoc methods can be also applied to the intrinsic model \cite{molnar_interpretable_2020}.  

We have added four new axes for categorizing and describing existing approaches: 
\begin{enumerate}[label=\textbf{A\arabic*}, leftmargin=*]
\setcounter{enumi}{2}
\item\label{enum:A3}time horizon of explanation (reactive/proactive),
\item\label{enum:A4}type of environment (deterministic/stochastic),
\item\label{enum:A5}type of policy(deterministic/stochastic),
\item\label{enum:A6}agent cardinality (single-agent/multi-agent).
\end{enumerate}

The first new axis in the taxonomy performs separation regarding the time-horizon of explanation.  With respect to that axis we have two groups:
\begin{enumerate}
    \item Reactive explanations where the explanation is focused on the immediate moment, considering only a very short horizon and momentary information. 
    \item Proactive explanations which focus on longer-term consequences. These explanations would consider information about the anticipated future, such as scenario rollouts or referring to some patterns in the anticipated future.
\end{enumerate}

We can say that a reactive explanation provides an answer for the “what happened” question, and proactive for the \textit{"why has something happened"} question. So, in the example above, a reactive explanation for the question: \textit{"Why did the car stop?"} would be: \textit{"A pedestrian appeared in the front of the car"}. A proactive explanation for the same questions would be: \textit{"If the car hadn't stopped, it would have run over a pedestrian. The pedestrian would suffer injuries that might have been fatal. The car didn’t want to injure the pedestrian and so it stopped."}. In this example, although the reactive explanation gave us a brief answer to the question "why did the car stop", the rest of the explanation unfolded in our head similarly to the proactive explanation. We can say that a reactive explanation in the presence of experience, transmitted or learned, generates a proactive explanation. 

Another example is strategy games such as chess and Go, where the current action (move) cannot always be explained so easily because the reason is not obvious and sometimes the usefulness of some action will become apparent in the future.
Here the explanation must be in the form of consideration to how the current move will affect the ability to perform a future move. It takes a lot more time steps compared to the current moment to explain one action and these methods are in this paper called proactive explanation methods.

The second new axis performs separation regarding the type of environment in the models being explained. Respect to this axis, there are two groups:
\begin{enumerate}
    \item Deterministic environment – an environment where the current state and action uniquely determine the next state.
    \item Stochastic environment – a random environment where the current state and action don’t uniquely determine the next state.
\end{enumerate}

The environment is a world that surrounds the agent. The agent stays within the environment in some situation, interacts with the environment and can change its current situation. That situation is called a state, and an interaction is called an action. If the agent in state $S1$ after action $A1$ always passes into state $S2$, then the environment is deterministic. If with some probability or random distribution the agent passes into state $S2$, the environment is stochastic.

The third new axis \ref{enum:A5} is classifies approaches according to the type of policy in the model being explained as follows:
\begin{enumerate}
    \item Deterministic policy – which consistently performs one of the available actions with a probability of 100\%, and
    \item Stochastic policy – which represents the function of a probability distribution that assigns a probability for each event. There is no action that will take place with certainty and different choices can be made in repeated situations.
\end{enumerate}

\begin{table*}[ht]
\centering
\caption{The main categorization groups of reviewed methods}
\label{tab:groups_subgroups}
\begin{tabular}{|M{0.01\linewidth}|M{0.3\linewidth}|M{0.3\linewidth}|M{0.3\linewidth}|}
\hline
                           & \textbf{ Group} & \textbf{Advantages} & \textbf{Disadvantages} \\ \hline
                           
\multirow{3}{*}{\rotatebox[origin=c]{90}{\parbox{1.7cm}{\normalsize Reactive}}}  &Policy simplification&The simpler model or policy  is inherently interpretable to human.&Reducing model performance and accuracy.\\ \cline{2-4} 
                           &Reward decomposition&Better understanding the cause of receiving the reward.&The limitation to the environments whose reward function has a natural decomposition into meaningful components \cite{juozapaitis_explainable_2019}.\\ \cline{2-4} 
                           &Feature contribution and visual methods&Frequent and easily applicable methods.&Only the current decision is explained. Certain domain knowledge is required.\\ \hline
                           
\multirow{4}{*}{\rotatebox[origin=c]{90}{\parbox{2cm}{\normalsize Proactive}}} &Structural causal model&People represent and see world through causal lens. This type of explanation is close to human way of thinking.&Inability to generate a real-time causal model which is finite and not predefined to certain problem.\\ \cline{2-4} 
                           &Explanation in terms of consequences&The ability of the user to query agent “why not” questions.&A user is expected to have an opinion. If he does not ask a good question, no explanation will be meaningful.\\ \cline{2-4} 
                           &Hierarchical policy&It is natural for humans to reuse previously learned skills as subcomponents of new skill \cite{shu_hierarchical_2017}. &Only the current goal is explained, not global policy. Tasks and subtasks are not automatically defined.\\ \cline{2-4} 
                           &Relational reinforcement learning&The ability to provide background knowledge by using a set of logical rules. &Complex problem with huge action space is hard to describe with relation language.\\ \hline

\end{tabular}
\end{table*}

The last new axis, \ref{enum:A6}, classifies reinforcement learning systems according to agent cardinality as follows:
\begin{enumerate}
    \item single-agent system – there is only one agent that interacts only with the environment, and
    \item multi-agent system – there is more than one agent and they interact with each other and with the environment.
\end{enumerate}
The challenge of multi-agent RL is finding the optimal policy for each agent with respect to both, environment and other agents’ policies \cite{nguyen_counterfactual_2021} Models and techniques used to simplify and explain RL policy are organized into two main groups depending on the type of explanation they can give: reactive or proactive. More precisely, they are divided into groups depending on whether one or more time steps are used to generate the explanation.
Each of these groups is further divided into subgroups depending on how the explainable policy is generated.

Reactive group:
\begin{itemize}
    \item Policy simplification:
    \begin{itemize}
        \item by tree structure
        \item by other methods
    \end{itemize}
    \item Reward decomposition
    \item Feature contribution and visual methods
\end{itemize}

Proactive group:
\begin{itemize}
    \item Structural causal model
    \item Explanation in terms of consequences
    \item Hierarchical policy
    \item Relational reinforcement learning
\end{itemize}

Table \ref{tab:groups_subgroups} lists the mentioned groups and subgroups together with their main advantages and disadvantages. 
Table \ref{tab:lit_summary}  presents the reviewed literature on XRL and its categorization according to the taxonomy described above. Additionally, each paper is described by additional columns depending on:
\begin{enumerate}
    \item scope of explanation,
    \item time of explanation,
    \item type of environment (deterministic /stochastic),
    \item type of policy (deterministic /stochastic),
    \item agent cardinality (single-agent/multi-agent),
    \item type of explanation (image, diagram, text).
\end{enumerate}

\newpage
\onecolumn

\begin{landscape}
\centering
\begin{longtable}{|M{0.02\linewidth}|M{0.05\linewidth}|M{0.35\linewidth}|M{0.05\linewidth}|M{0.07\linewidth}|M{0.08\linewidth}|M{0.08\linewidth}|M{0.1\linewidth}|M{0.05\linewidth}|}
    
    \caption{Summary of reviewed literature in the field of explainable reinforcement learning}
    \label{tab:lit_summary}\\

\hline
 & & Paper, year & Scope
 
 (Global / Local) & Time 
 
 (Post hoc / Intrinsic) & Environment 
 
 (det./ stoch.) & Policy 
 (det./ stoch.) & Type of explanation & Agent cardinality \\ \hline 
 
\multirow{15}{*}{{\rotatebox[origin=c]{90}{\parbox{14cm} {\LARGE \centering Reactive}}}} & \multirow{6}{*}{\rotatebox[origin=c]{90}{\parbox{5cm}{\large\centering Policy simplification}}} & Particle Swarm Optimization for Generating Interpretable Fuzzy Reinforcement Learning Policies, 2017. \cite{hein_particle_2017}  & Global & Intrinsic & Stochastic & Stochastic & Image (diagrams), IF-THEN rules & Single-agent \\ \cline{3-9} 
 & & Toward Interpretable Deep Reinforcement Learning with Linear Model U-Trees, 2019. \cite{liu_toward_2019}  & Global and local & Post hoc & Stochastic & Stochastic & Graph. image & Single-agent \\ \cline{3-9} 
 & & Distilling Deep Reinforcement Learning Policies in Soft Decision Trees, 2019. \cite{coppens_distilling_2019}  & Global & Post hoc & Stochastic & Stochastic & Image (heatmap) & Single-agent \\ \cline{3-9} 
 & & Conservative Q-Improvement: Reinforcement Learning for an Interpretable Decision-Tree Policy, 2019.  \cite{roth_conservative_2019} & Global & Intrinsic & Stochastic & Deterministic & Graph & Single-agent \\ \cline{3-9} 
 & & Programmatically Interpretable Reinforcement Learning, 2019. \cite{verma_l_2019}  & Global & Intrinsic & Stochastic & Deterministic & high-level, domain-specific prog. lang. & Multi-agent \\ \cline{3-9} 
 & & Interpretable policies for reinforcement learning by empirical fuzzy sets, 2020. \cite{huang_interpretable_2020}  & Global & Intrinsic & Stochastic & Stochastic & IF-THEN rules & Single agent \\ \cline{2-9} 
 & {\rotatebox[origin=c]{90}{\parbox{2cm}{\large \centering Reward 
 
 decomp.}}} & Explainable Reinforcement Learning via Reward Decomposition, 2019. \cite{juozapaitis_explainable_2019}  & Local & Post hoc & Stochastic & Deterministic & Image (diagrams) & Single-agent \\ \cline{2-9} 
 & \multirow{8}{*}{\rotatebox[origin=c]{90}{\parbox{7cm}{\large \centering Feature contribution and visual methods}}} & Graying the black box: Understanding DQNs, 2016. \cite{zahavy_graying_2016} & Global & Post hoc & Stochastic & Deterministic & Image & Single-agent \\ \cline{3-9} 
 & & Visualizing and Understanding Atari Agents, 2018. \cite{greydanus_visualizing_2018}  & Global & Post hoc & Deterministic & Deterministic & Image (Saliency map) & Multi-agent \\ \cline{3-9} 
 & & Reinforcement Learning with Explainability for Traffic Signal Control, 2019. \cite{rizzo_reinforcement_2019}  & Local & Post hoc & Deterministic & Stochastic & Image (charts) & Single-agent \\ \cline{3-9} 
 & & RL-LIM: Reinforcement Learning-Based Locally Interpretable Modeling, 2019. \cite{yoon_rl-lim_2019}  & Local & Post hoc & Deterministic & Deterministic & Image (saliency maps) & Single-agent \\ \cline{3-9} 
 & & Self-Supervised Discovering of Interpretable Features for Reinforcement Learning, 2021. \cite{shi_self-supervised_2021} & Global & Post-hoc & Stochastic & Stochastic & Image & Single-agent \\ \cline{3-9} 
 & & Interpretable End-to-End Urban Autonomous Driving With Latent Deep Reinforcement Learning, 2020. \cite{chen_interpretable_2020} & Local & Post-hoc & Stochastic & Stochastic & Image & Multi-agent \\ \cline{3-9} 
 & & Explainable Deep Reinforcement Learning for UAV Autonomous Navigation, 2021.  \cite{he_explainable_2021}  & Global & Post hoc & Stochastic & Stochastic & Image (saliency map, diagrams), text & Single-agent \\ \cline{3-9} 
 & & Counterfactual Explanation with Multi-Agent Reinforcement Learning for Drug Target Prediction, 2021.  \cite{nguyen_counterfactual_2021} & Local & Post hoc & Stochastic & Stochastic & Image (graph- based) & Multi-agent \\ \hline
 
 \multirow{13}{*}{\rotatebox[origin=c]{90}{\parbox{12cm}{\LARGE \centering Proactive}}} & \multirow{3}{*}{\rotatebox[origin=c]{90}{\parbox{2cm}{\large\centering Structural causal model}}} & Explainable Reinforcement Learning Through a Causal Lens, 2020.  \cite{madumal_explainable_2020}  & Local & Post hoc & Stochastic & Stochastic & Diagram, text & Multi-agent \\ \cline{3-9} 
 & & Distal Explanations for Model-free Explainable Reinforcement Learning, 2020. \cite{madumal_distal_2020}  & Local & Intrinsic & Stochastic & Stochastic & Diagram & Multi-agent \\ \cline{3-9} 
 & & Deep Structural Causal Models for Tractable Counterfactual Inference, 2020.  \cite{pawlowski_deep_2020} & Global & Post hoc & Stochastic & Stochastic & Diagram, image & Single-agent \\ \cline{2-9} 
 & \multirow{6}{*}{\rotatebox[origin=c]{90}{\parbox{6cm}{\large\centering Explanation in terms  of consequences}}} & Improving Robot Controller Transparency Through Autonomous Policy Explanation, 2017. \cite{hayes_improving_2017} & Global & Post hoc & Stochastic & Stochastic & Text & Single-agent \\ \cline{3-9} 
 & & Contrastive Explanations for Reinforcement Learning in terms of Expected Consequences, 2018. \cite{van_der_waa_contrastive_2018} & Local & Post hoc & Stochastic & Stochastic & Text & Single-agent \\[1ex] \cline{3-9} 
 & & Feature-Based Interpretable Reinforcement Learning based on State-Transition Models, 2021. \cite{davoodi_feature-based_2021} & Local & Post hoc & Deterministic & Stochastic & Text & Single-agent \\[1ex] \cline{3-9} 
 & & Tracking Predictive Gantt Chart for Proactive Rescheduling in Stochastic Resource Constrained Project Scheduling, 2018 \cite{brcic_tracking_2018}  & Global & Post hoc & Stochastic & Deterministic & Diagram & Single-agent \\[1ex] \cline{3-9} 
 & & Planning horizons based proactive rescheduling for stochastic resource-constrained project scheduling problems, 2018 \cite{brcic_planning_2019} & Global & Post hoc & Stochastic & Deterministic & Diagram & Single-agent \\[1ex] \cline{3-9} 
 & &EDGE: Explaining Deep Reinforcement Learning Policies, 2021 \cite{guo2021edge}  & Global & Post hoc & Stochastic  & Stochastic & Image & Multi-agent \\[1ex] \cline{2-9} 
 & \multirow{2}{*}{\rotatebox[origin=c]{90}{\parbox{2cm}{\large\centering Hierarchical policy}}} &  Hierarchical and Interpretable Skill Acquisition in Multi-task Reinforcement Learning, 2017. \cite{shu_hierarchical_2017}  & Local & Intrinsic & Stochastic & Stochastic & Text  & Single-agent \\[4ex] \cline{3-9} 
 & & Dot-to-Dot: Explainable Hierarchical Reinforcement Learning for Robotic Manipulation, 2019. \cite{beyret_dot--dot_2019}   & Global & Intrinsic & Stochastic & Deterministic & Image (heatmap) & Single-agent \\[4ex] \cline{2-9} 
 & \multirow{2}{*}{\rotatebox[origin=c]{90}{\parbox{3cm}{\large\centering Relational reinforcement learning}}} & Relational Deep Reinforcement Learning, 2018. \cite{zambaldi_relational_2018} & Local & Intrinsic & Stochastic & Stochastic & Images & Multi-agent \\[6ex] \cline{3-9} 
 & & Learning Symbolic Rules for Interpretable Deep Reinforcement Learning, 2021. \cite{ma_learning_2021}  & Global & Intrinsic & Stochastic & Stochastic & Text (rules . First Order Logic) & Single-agent \\[6ex] \hline
\end{longtable}

\end{landscape}

\twocolumn

\subsection{Reactive explanation }
\subsubsection{Policy simplification}

In deep reinforcement learning (DRL) deep neural networks are used to approximate functions that are used in traditional Reinforcement learning. Different DRL algorithms use DNNs in different ways: for policy optimization in policy gradient RL \cite{sutton_policy_1999}, to approximate the value function or Q-function or to approximate the transition function. Since the way the neural network works is hard to understand and interpret, it is logical that the same applies to the policy it generates.

One group of the approaches described in this group attempted to learn policy into a form that is interpretable by itself. Example of used models for this purpose are decision trees and fuzzy rulesets. Another group aimed to distill the learned policy into interpretable forms such as trees or domain-specific programming languages. Their policy can be easily presented in text or graphics and thus be understandable to people. The disadvantage of the first group approach is the sacrifice of the initial model performance by reducing complexity to achieve explainability. The more understandable the model, the lower its accuracy. By distilling policy this problem is solved, but as the complexity of the model increases, such explanations begin to lose their usefulness. 

\textit{Policy simplification with tree structure}

Liu et al. \cite{liu_toward_2019} introduced Linear Model U-trees (LMUTs) to approximate a Q-value function learned by a deep neural net. They extend the classic U-tree, classic RL algorithm which represents a Q-function using a tree structure, by adding a linear model to each leaf node. LMUT used mimic learning where learner observed interaction between neural net and environment. LMUT enables interpretability by analyzing feature influence, extracting rules, and highlighting pixels in image input \cite{liu_toward_2019}. 

Coppens et al. \cite{coppens_distilling_2019} make the RL policy more interpretably using a Soft Decision Tree (SFT). In a classic binary tree, each node according to the applied test and the obtained outcome, choose one of the branches. And this process is repeated from the root to a leaf which contains the final result/label/class. A soft decision tree is also a binary tree, but in its decision node both branches (children) are chosen, but with some probabilities. Precisely, each decision node consist of a single perceptron, with a weight and bias parameter, and for a given input $x$ the probability for choosing right or left child is determined \cite{coppens_distilling_2019}. To assign input data to a particular leaf with a certain probability, the model learns a hierarchy of filters in its decision (branching) nodes \cite{coppens_distilling_2019}. Compared to the work above, this work focuses on mimicking an explicit policy. After training, they distill the network policy to a soft decision tree and use heatmap to visualize learned filters. They tested this method on the Super Mario gamer and the goal was to get insights about the way DRL agent controls Mario actions. More interesting area for next Mario’s move was highlighted red on the heat map. But this kind of explanation can only give a momentary explanation of why Mario jumped there - because these pixels were the reddest.
Roth et al. \cite{roth_conservative_2019}  suggest a slightly different method for achieving interpretability of RL policy using trees. They propose a novel algorithm, the Conservative Q-Improvement (CQI) reinforcement learning algorithm, which learns a policy in the form of a decision tree \cite{roth_conservative_2019}. Unlike the previously described model that after training distills the output of the deep policy to a soft decision tree and analyzes it, CQI maintains the policy as a tree at all stages of training and updates the tree. This algorithm uses a lookahead approach that predicts which split will increase reward the most and only in this case will the tree size be increased. 

\textit{Another way of simplifications RL policy}

Verna et al. \cite{verma_l_2019} proposed another way to simplify policies discovered by DRL. They present a reinforcement learning framework Programmatically Interpretable Reinforcement Learning (PIRL) which represents policies using high-level, domain-specific programming language which is human-readable. Since the space of policies permitted can be vast, they also proposed a new optimization method for finding the programmatic policy with the greatest reward, Neurally Directed Program Search (NDPS).

Hein et al. presented in \cite{hein_particle_2017} simplification RL policy using FUZZI controllers, interpretable system controllers for continuous state and action space. They proposed a fuzzy particle swarm reinforcement learning (FPSRL) approach that can solve problems in domains where online learning is forbidden and which is human-readable and understandable. A Fuzzy controller is specified by a set of linguistic IF-THEN rules that can be easily visualized and interpreted and whose membership function can be activated independently and produce a combined output \cite{hein_particle_2017}. While this method doesn’t learn policies online, in a per-step manner, if something changes in the environment, the policy parameters will not be updated until the end of an episode and the whole set of policy parameters has to be relearned \cite{huang_interpretable_2020}. Also, the number of rules must be given a priori. Huang et al. \cite{huang_interpretable_2020} proposed interpretable fuzzy reinforcement learning (IFRL) method that learns online in a step by step manner and is able to react to the change of the environment in real time. The learned policy can be expressed as human-understandable IF-THEN rules. There is still a lot of room for improvement in this method like reducing number of rules, better balance between performances and the number of nodes an so on. 

\subsubsection{Reward decomposition}

There are efforts to achieve explainability by decomposing rewards into meaningful components \cite{juozapaitis_explainable_2019}.  This is also the disadvantage of this approach because it is limited to the environments whose reward function has a natural decomposition into meaningful components. For example in the Dota 2 RL domain some reward types are: kills, assists and so on, and each type brings a certain size of the reward \cite{juozapaitis_explainable_2019}.
In a typical RL all individual rewards are summed up and total Q-values are computed as a total amount. In this way, it is not possible to gain insight into the positive and negative factors that contribute to decision-making. Here, the Q-function is decomposed into reward types. In \cite{juozapaitis_explainable_2019} they are trying to explain why one action is preferred to another in some state in terms of meaningful reward types.  This way the policy can be better understood, i.e. whether the agent is doing something to be closer to the goal or to avoid penalties \cite{juozapaitis_explainable_2019}. In both cases, the agent will receive an award, and this helps the human to understand the reason for receiving the award.

\subsubsection{Feature contribution and visual methods}

The aim of the proposed methods in this group is to extract the features that contribute the most to decision making and use them to generate explanations. Some of the main methods for feature extraction are: Local Interpretable Model Agnostic Explanations (LIME), Shapley values, DeepLIFT, Grad CAM. The disadvantage of all these approaches is that only the current decision is explained, but the whole decision-making process remains unexplained. In a word, there is a lack of proactivity.
Saliency maps are one of the most well-known visual methods where the intensity of the pixel color at a particular location corresponds to the importance of the input value. Greydanus et al. in \cite{greydanus_visualizing_2018} introduced a method for generating useful saliency maps in order to understand how an agent learns and executes a policy. They divided saliency maps into two groups: gradient based and perturbation-based saliency methods. The first group tried to show what features of DNN’s input have the biggest impact on the output, while the second one measured the sensitivity of the output to an alteration in the input. They analyzed Atari RL agents that use raw visual input to make their decision. 
They focus on agents trained via the Asynchronous Advantage Actor Critic  (A3C) algorithm. So, their goal was to visualize maps for the actor policy $\pi$ and a value (critic) $V$$\pi$ for each step, to identify key information in the image that the policy uses to select an action and information for assigning a value at time $t$.

In \cite{rizzo_reinforcement_2019} values are used to generate a reactive explanation. SHAP values are based on Shapely values from the game theory concept. Compared to game theory, where the contribution of each player is measured, the feature value data act as players here, and the idea is to compute the contribution of each feature to the prediction \cite{molnar_interpretable_2020}.  In \cite{rizzo_reinforcement_2019} the authors dealt with a traffic light control optimization problem. They trained a neural network to select the traffic light phase and represent the model as a MDP where the state is a count of cars detected from the detector in a time step, an action is the selection of the traffic lights, and the reward is the number of vehicles departed from the roundabout in the allotted time. By computing Shapley values of each detector, they showed that the influence of some detectors was positive or negative to the choice of traffic light phase selection.

He et al. in \cite{he_explainable_2021} also used the SHAP method to provide decisioned explanations in UAV (Aerial Vehicles) obstacle avoidance. Precisely, they used Deep SHAP that is the connection between Shapley values and DeepLIFT. DeepLIFT provides an approximation of Shapely values by using the Rescale rule and RevealCancel rule. Otherwise, when a model is fully linear, the SHAP values are obtained by summing the attributions along all possible paths between input and output \cite{he_explainable_2021}. But in their network, there are non-linear parts, and that is why DeepLIFT was used. To visualize the CNN perceptron, they proposed a new saliency method that combined both CAM and SHAP values. This method is called SHAP-RAM (SHAP value-based regression activation map) because their problem is a regression problem. Their method can be used in any network architecture with a global average pooling layer which purpose is to summarize the activation of the last CNN layer \cite{he_explainable_2021}.  Because trained data is not perfect, there are explanations that don't make much sense. 

Yoon et al. \cite{yoon_rl-lim_2019} addressed the challenges in the representational capacity difference while applying distillation. More specifically, black-box models, like DNN, have a much larger representational capacity than locally interpretable models. Their goal is to make a locally interpretable model whose predictions are similar to the predictions of the pre-trained black-box model. To achieve this while avoiding overfitting, they want to find a small number of valuable instances and use them to train a low-capacity locally interpretable model \cite{yoon_rl-lim_2019}. So, they proposed the RL-LIM (Reinforcement Learning-Based Locally Interpretable Modeling) method, which used reinforcement learning to integrate the reward directly with the fidelity metric to find an optimal policy that maximizes the fidelity of the locally interpretable model. The greater similarity between the models, the greater the reward is obtained.

Shi et al. \cite{shi_self-supervised_2021} propose another framework for discovering interpretable features – the Self-Supervised Interpretable Network (SSINet). This framework is applicable to an RL model taking images as input. It produces a fine-grained attention mask for highlighting task-relevant information which gives the best explanation for made decisions. 

Chen et al. \cite{chen_interpretable_2020} proposed a new interpretable RL method for end-to-end autonomous driving car. The driving policy gets camera and lidar images as input and generates driving commands. This policy is learned along with a sequential latent environment model that can generate a bird’s-eye-view semantic mask and provide an explanation of how the car sees and perceives the driving situation. From the generated bird-view mask it is possible to see which surrounding cars the autonomous vehicle has located and decode the map information even they are not directly form the sensor. The disadvantage of this approach is that the process of decision-making itself has remained unexplained.

In \cite{zahavy_graying_2016} they present a method to analyze Deep Q-networks. The big problem in RL is high-dimensional data like speech input. They used t-SNE for visualizing high-dimensional data which uses a t-student distribution that enables the visualization of different sub-manifolds learned by the network and interpretation of their meaning. They apply t-SNE directly on the collected neural activations. Different features are present in each sub-manifold. They use a 3D t-SNE state representation for visualizing the state transitions of the learned policy that reveal hierarchical structures. They also proposed the Semi Aggregated Markov Decision Process (SAMDP) that combines advantages of the SMDP and AMDP approaches. These two methods are extensions of the MDP model. SAMDP allows the identification of spatio-temporal abstractions directly from the learned representation. They show that they can offer an interpretation of learner policies and their main contributions are understanding, interpretability and debugging. 

Nguyen et al. \cite{nguyen_counterfactual_2021}  proposed a Multi-Agent Counterfactual Drug-target binding Affinity (MACDA) framework, which uses multi-agent reinforcement learning for explaining deep models of drug-target binding affinity. To achieve explanation, they use a counterfactual type of explanation. The goal is to find an important part of the input by making small changes in the input that will result in large changes in the output. This framework consists of two main components: the agent and the environment. The agent chooses a modified version of the input drug and protein, then the model of the environment receives them and calculates the binding affinity between the drug and protein. Just some chemical and biological processes are taken into account in this paper so it opens up opportunities for future research.

\subsection{Proactive explanation}
\subsubsection{Structural causal model}

One approach of achieving explainability in RL is learning a structural causal model during reinforcement learning with the aim of understanding the relationships between variables. This way of generating a explanation is very close to humans, because we view the world through a causal lens \cite{sloman_causal_2005}. The weakness of this approach is the inability to generate a real-time causal model which is finite and not predefined to certain problem.

\noindent\textbf{DEFINITION OF THE CAUSAL MODEL}

According to the Cambridge dictionary, causal means a relationship or link between two things in which once causes the other.

Schölkopf \cite{scholkopf_causality_2019} says that the hard open challenges of ML and AI are essentially related to causality, including answering counterfactual questions. The causal model describes the world by random variables \cite{halpern_causes_2013}. Random variables have a causal influence on each other and that influence is modeled by a set of structural equations. These variables are divided into two sets:
\begin{itemize}
    \item \textit{exogenous} (external) variables whose values are determined by outside factors and 
    \item \textit{endogenous} (internal) variables whose values are determined by the structural equations.
\end{itemize}

Formally, a signature $S$ is a tuple $(U, V, R)$ where $U$ is a finite set of exogenous variables, $V$ is a finite set of endogenous variables, and $R$ is a function that associates possible values from nonempty set $R(Y)$ for every variable $Y \in (U \cup V)$ \cite{halpern_causes_2013}.

\begin{definition} A structural \textit{causal model} (\textit{SCM}) over signature S is a tuple $M = (S,F)$, where: $F$ associates with each variable $X \in V$ a function denoted $F_x$ such that $F_{x} : (x_{u \in U} R(u)) \times (X_{Y \in V-\{x\}}R(Y)) \rightarrow R(X)$. 
$F_{x}$ tells us the value of $X$ given the values of all the other variables in $U \cup V$ \cite{halpern_causes_2013}.\end{definition}

Examples of endogenous variables are: fire (F), lightning (L) or match lit (ML). Exogenous variables are things we need to assume that make a forest fire possible like: wood is sufficiently dry, there is enough oxygen in the air, etc.  If $\overrightarrow{u}$ is a vector of values of each exogenous variable $u \in U$ then, for example 
$F_F (\overrightarrow{u}, L, ML)$ is such that $F = 1$ (the fire happened) if $L = 1$ or $ML = 1$ \cite{halpern_causes_2013}.. 

\noindent\textbf{PAPERS BASED ON A CAUSAL MODEL}

Madumal et al. \cite{madumal_explainable_2020} present a model that generates explanations of behavior based on counterfactual analysis of the \textit{structural causal model} that is learned during reinforcement learning. 
In \cite{madumal_explainable_2020} the given definition of the \textit{causal model} is extended to include actions as a part of the causal relationship. With the addition of actions, they incorporate action influence models for model-free reinforcement agents based on MDP. They aim in this paper to explain an agent's behavior using their understanding of how actions affect the environment. The authors defined a signature $S_a$ that defines tuple $(U, V, R, A)$, in which $U$, $V$ and $R$ are same as above, and $A$ is a set of actions. 

\begin{definition} An \textit{action influence model} is a tuple $(S_a, F)$, where $S_a$ is as above, and $F$ is the set of structural equations. There are multiple structural equations for each $X \in V$, one for each unique action set that influences $X$. A function $F_{x.\textsc{A}}$, for $ \textsc{A} \in A$, defines the causal effect on $X$ from applying action $\textsc{A}$ \cite{madumal_explainable_2020} . \end{definition}

After the qualitative causal relationships of variables are defined as an action influence model and the structural equations during RL are learned, the third phase is generating explanans from SCM \cite{madumal_explainable_2020}. A causal chain is a path between a set of events, where a path from event $X$ to event $Y$ indicates that $X$ has to occur before $Y$. They \cite{madumal_explainable_2020} defined a complete explanans for action $A$ as the complete causal chain from action $A$ to any future reward that it can receive. That chain includes the first and last arcs, with their source and destination node, with all intermediate nodes. Here a problem with large graphs appears, where the complete causal chain can be too long, so they also defined minimally complete explanation. For minimal complete causal chain intermediate nodes are omitted. 

They also defined explanations for questions ‘Why action A’ and ‘Why not action B’. For why question they used this model to define a minimally complete explanation, and for “Why not”  questions they also defined a minimally complete contrastive explanation \cite{madumal_explainable_2020}. The contrastive explanation is generated by finding the differences between extraction of the actual causal chain for action $A$, and the causal chain for counterfactual action $B$.

One weakness of this approach is that the causal model must be given in advance. Causal explanations from the action influence model proved to perform better than the state-action base explanation model, but the problem is in the use of structural equations that gives poor results in computational task prediction accuracy \cite{madumal_distal_2020}.
 
Madumal et al. \cite{madumal_distal_2020} introduce distal explanation model that uses a decision tree to generate explanations from causal chains, instead of structural equations that approximate the causal effect between two features. They defined distal action in RL as an action that depends mostly on the execution of the current action. To perform a distal action, the agent must first execute some other action. They also defined minimally complete distal explanations for “why” and “why not” questions. 

The weakness of this model stems from the fact the causal graph needs to be faithful to the problem, in order to learn the opportunity chains \cite{madumal_distal_2020}. The causal model used in this paper is a handcrafted model. The problems that occur with large domains and automatic real-time causal model generation remain.
Pawlowski et al.\cite{pawlowski_deep_2020} introduced a novel general framework for building structural causal models and demonstrated it with two case studies: a synthetic task and a real world example with brain MRI. 

Herlau in \cite{herlau_causal_2020} considers the problem of building a general RL agent which uses experience to construct a causal graph of the environment and then explains the policy with that graph. They showed a method that can learn a possible causal graph in a grid-world environment. A causal graph is an equivalence class of structural models \cite{herlau_causal_2020} wherein each vertex corresponds to the certain environment variable. The Bellman equation is used to estimate expectation and it is obtained by recursively decomposing the value function \cite{herlau_causal_2020}. Bellman equation takes into account that the usefulness of states depends not only on the immediate “reward” but also on future reduced rewards. But the construction of a full causal graph remains an open problem. 

\subsubsection{Explanation in terms of consequences}

Van der Waa et al. in \cite{van_der_waa_contrastive_2018} made a study that showed that human users tend to favor explanations about policy rather than about a single action. So they propose an approach to XRL, based on the Markov Decision Problem (MDP), that allows an agent to answer a question about its policy in terms of consequences. This allows a human user insight to what the agent can perceive from a state and which outcomes it expects from an action or visited state. The disadvantage of this approach is that it generates an explanation only when humans have an opinion or doubts that something else should have been done instead of the already performed action \cite{hayes_improving_2017, van_der_waa_contrastive_2018}. Also, how an individual action will affect a situation is problem-specific. A model should be created beforehand and, once created, it is used to explain any agent’s interactions \cite{davoodi_feature-based_2021}. For complex and unbounded environments it is not possible to explore possible transitions between states. Hayes and Shah \cite{hayes_improving_2017} first developed a method that can generate an explanation about learned policy that is grounded in language that humans can understand. In their method, it is first necessary to define the list of predicates, encoded as binary classifiers over the state space, e.g. ”looking high”, ”part detected”, ”at delivery area” and so on. When some action is performed, the subset of predicates on the list that have a true value are correlated with that action. So these predicates are used to answer behavior-related questions like “When do you do \textunderscore?”,  “Why did not you do \textunderscore?” or “What you do when \textunderscore?”.  This type of explanation is good for small problems, but when problems are bigger and more complicated, the list of predicates become too large and the explanation does not make sense anymore. Also, the disadvantage of this method is that it does not give any explanation for why an agent did some action. 
To limit the amount of information of all consequences, \cite{van_der_waa_contrastive_2018} proposed a method that supports contrastive explanations.  Their method allows the user to formulate a question in “why” format: “why something happened instead of something else”. They introduced two policies, the learned policy  $\pi_t$ (the “fact”) and policy in of interest to the user $\pi_f$ (the foil). So, the previous question can be formulated as follows: “why the learned policy $\pi_t$ instead of $\pi_f$”. To answer these questions, two simulations are performed, one constructs a Markov Chain under policy $\pi_f$ and the under policy $\pi_t$. The difference between these two simulations can be presented as an answer. Instead of giving explanations in terms of rewards (numeric values), they developed a framework that translates states and actions into a description that is easier to understand for human users. This is similar approach the one from \cite{hayes_improving_2017}.
The original set of states can be transformed to a descriptive set $C$ according to the function $k: S \rightarrow C$. A similar thing is done with the set of rewards that are explained in terms of action outcome $O$, according to the function: $C \times A \rightarrow Pr(O)$ where $Pr(O)$ is a probability distribution over $O$. A classic MDP tuple is $(S,A,R,T, \lambda)$ where $S$ is the set of states, $A$ is the set of actions, $R$ is the reward function $R: S \times A \rightarrow \mathbb{R}$  and $T$ is the transition function $T: S \times A \rightarrow Pr(s)$ and $\lambda$ is the discount factor. This tuple is extended to the new MDP tuple  $(S, A, R, T, \lambda, C, O, t, k)$ where $C$ and $O$ are descriptive sets and $k$ and $t$ functions \cite{van_der_waa_contrastive_2018}.

Davoodi and Komeili \cite{davoodi_feature-based_2021} consider feature-based interpretability in order to explain the risk by doing some action in the current state. They found that a human should be able to visualize action sequences that change the value of some feature that is associated with increased risk. They used a state transition graph to determine the state transition model. Each node represents a set of states and the edge between two nodes represents the fact that exists at least one action the leads from one state in first node to the state in second node. After the graph is constructed, the goal is to find all risky nodes. The node is considered as risky if all its outgoing edges lead to risky states \cite{davoodi_feature-based_2021}.
Instead of generating an explanation based on the features of a particular state that contributed the most to a particular action and thus getting a certain reward, Guo et al. in \cite{guo2021edge} propose a different method. Their method Strategy-level Explanation of Drl aGEnts (EDGE) identifies the critical time steps within each episode that contribute to a agent's final reward. They achieved this by extending the Gaussian process (GP) with a customized deep additive kernel that captures the correlations between time steps and their joint effects across episodes. Also, they design an interpretable Bayesian prediction model to predict the final reward and extract the time step importance. This type of explanation makes it possible to understand what are the crucial steps that are important for agent’s action and what are the consequences of taking/not taking optimal actions at those steps. With such an explanation one can better understand the overall agent policy and find policy vulnerabilities.

In \cite{brcic_tracking_2018} an extension of predictive Gantt chart is presented. It can track the execution of the stochastic project through the time and using  simulation statistics generate visual presentations of the consequences of using current policy in the future. So it can possibly show that behavior of current methods does not match the real-world intuition. 

\subsubsection{Hierarchical policy}
The idea of hierarchical policy is to decompose one task into multiple subtasks that are at different abstraction levels. A prerequisite for performing a task in order to perform the next one is used as an explanation of a particular action. To achieve interpretability, each task is described by a human \cite{shu_hierarchical_2017}. The disadvantage of the described methods is the possibility of explaining only the current agent’s goal, and not the global policy and wider picture. Also, a human is required to define the skill to be learned in each training stage \cite{shu_hierarchical_2017}.
There are two sets of policies in hierarchical RL: 
\begin{itemize}
    \item local policies – use primitive actions for achieving subtasks, 
    \item global policies – manage local policies in a sequence.
\end{itemize}

Local policies, also defined in \cite{kulkarni_hierarchical_2016} as lower-level policies, are trained in general RL algorithms to map states to primitive actions. Global policies, called ‘meta policies’\cite{kulkarni_hierarchical_2016}, are trained to select which of these lower-level policies to apply over a trajectory in order to start the sequence of subgoals to achieve the final goal. In \cite{shu_hierarchical_2017}, unlike previous works that used a two-layer hierarchical policy, they aim to learn multi-level global policy which means that global tasks can used lower-level global tasks as sub-tasks. 

In another paper \cite{shu_hierarchical_2017}, the stochastic temporal grammar (STG) was used to summarize temporal transitions between various tasks which is learned via self supervision. Knowing the order of the tasks, an explanation for one subtask can be given as the reason and condition for performing the next task. For example, to move some object from one place to another, one first needs to find that object, pick it up and put it down. Another good side of the hierarchical policy is the ability to reuse previously learned skills alongside and as subcomponents of new skills \cite{shu_hierarchical_2017}. For achieving this, discovering the relations between skills is needed. Applying Hierarchical RL is a very natural choice for clinical applications so interpretability of this policy is very important \cite{liu_deep_2019}. The framework described in this paper \cite{shu_hierarchical_2017} needed humans to define what skills to be learned in each training stage. 

Another attempt to use hierarchical policy in deep reinforcement learning and achieve explainability in the robotic system is \cite{beyret_dot--dot_2019}. The goal is to achieve human realization of the way in which robots interpret human actions. This is for the purpose of achieving trust and safety. There are two agents, a high-level agent that will divide the full task into smaller actions (sub-goals) for a low-level agent, which follows the tasks one by one \cite{beyret_dot--dot_2019}. They tried to achieve explainability  by using a heatmap on which the subgoals with higher $Q$ values are marked. The agent attributes higher values to sub-goals close to the end-goal rather than those closer to the starting position. This can only show that the agent learned a good representation of its environment, but it cannot give an exact explanation as to why the robot took some action at a given point in time. It offers a sense of security and trust to humans that the robot has learned the policy well. 

\subsubsection{Relational reinforcement learning}

In relation to reinforcement learning an action, states and policies are presented by relation language,  so it can be said that relation reinforcement learning is a combination of RL with relational learning or inductive logic programming \cite{zambaldi_relational_2018}. Relational representations facilitate the use of background knowledge in a natural way, provided by logical rules, and this has been known for a very long time in inductive logic programming (ILP) \cite{otterlo_survey_2005}. Block world is an example, where the goal is to stack one box $(A)$ on the top of another box $(B)$ that is on the floor, state S1 can be presented by a list of facts: $s1 = {clear (blockA), on (blockA, blockB) , on (blockB, floor)}$ \cite{dzeroski_relational_2001}. When the problems are simpler, it is possible to discover the relations between the states and there may be some interesting generalization capabilities \cite{zambaldi_relational_2018}. But with more complex problems when action space is not finite and combinatorial problems arise the generalization is much harder to achieve. 

The problem with RL is that it still doesn't seem to recognize when small differences occur in a task. 
Relational reinforcement learning allows the reuse of a simpler domain learning result in a more complex domain \cite{dzeroski_relational_2001}. So, when an agent learns how to move 3 blocks, it can move 4 blocks as well.
In this work \cite{dzeroski_relational_2001} the authors take into account pairwise non-local interaction between agents and show that these interactions, using a shared function, will be better studied for learning relations than an agent that consider only local interaction.  Policies in relational reinforcement learning are mostly represented as relational regression trees \cite{das_fitted_2020} which are more interpretable then NN.

Ma et al. \cite{ma_learning_2021} propose a Neural Symbolic Reinforcement Learning framework that consists of three components: a reasoning module, an attention module and a policy module. The reasoning module is based on a neural attention network and learns the logical rules in a symbolic rule space. The attention module can generate weights on predicates at different reasoning steps according to the symbolic information of the current RL state. Then, the policy module finally generates a DRL policy. So, interpretability is achieved by extracting the logical rules learned by the reasoning module \cite{ma_learning_2021}.
The advantage of this approach is memory conservation, because instead of storing all the rules in memory, this framework can extract rules selected by the attention modules. These rules are chain-like logical rules that represent relational paths in a knowledge graph, where objects are nodes and relations are edges. There are multiple paths from one node to another, which means that one action can be described with multiple rules. The disadvantage of this approach is that the extracted rules are not always true, so humans need to select one. Also, they only consider a finite action space.

\section{Proposal}
\label{sec:prop}
Explanations can be considered in three basic ways: 
\begin{itemize}
    \item as a tool for agreements in cooperation,
    \item as a tool for deception in adversarial situations,
    \item as a combination of both in mixed cooperative-competitive situations.
\end{itemize}

Unfairness of explainability \cite{brcic_impossibility_2022} shows that under certain conditions an explainer has a strategic advantage and can serve plausible explanations for potentially harmful decisions. In such cases, misinformed consent can be obtained, for potentially bad social outcomes. 
The explainer does not need to be an AI, but it can also be some human proxy that uses an AI decision-making model for their own benefit. For example, a bank might use a credit granting model that uses illegally obtained information. Or the model might be discriminating against some groups in a way to avoid detection by simple means usually employed in auditing and forensics. In that case, a bad decision might be made against a customer, for illegal reasons. But, customers might be unaware of that and their rights to contest the decision through courts if the bank delivers a plausible explanation consistent with the information available to the customer.

We aim to set standards for explanation that could hedge the worst case in adversarial situations to promote a more ethical behavior and wide-scale cooperation. We envision these standards would help auditing tools by making it harder for malicious explainers to trick validators. In the case of honesty on both sides, better understanding could be achieved through such explanations. Ideally, understanding of better explanations would map more faithfully to decision outcomes.

Our position for truthful explanations has three pillars:
\begin{enumerate}[label=\textbf{P\arabic*}, leftmargin=*]
 \item \label{enum:P1}We hold that explanations should be \textbf{proactive} in order to inform of potential profiles of consequences. As already explained, reactivity is just a special case of proactivity. Reactivity is usually invoked when there is an intuitive well-known environment such that full proactivity is implied and understood by human agents, or the time-horizon entailed by the problem is myopic. Proactivity demands some kind of anticipative model: scenario-trees, black-box sampler, or white-box model.
\item \label{enum:P2}	Explanations, just as the solutions, must take into account \textbf{risk attitudes} of involved parties, not just act in risk-neutral way as most of the research does.
\item \label{enum:P3}	The decision-maker’s computational process must be taken into account in a form of \textbf{epistemological constraints} due to the different limitations of resources, available information and algorithmic-procedural limitations with respect to efficiency and trade-offs which, in addition to the cooperative stance of the agent, affect its beliefs about decisions and explanations.
\end{enumerate}

\noindent\textbf{Proactivity of explanations (\ref{enum:P1}) }

People are intrinsically interested in the consequences, or in absence of perfect information, they would like to know the worst, best and likely consequences. As mentioned, reactive explanations such as in supervised learning and even in explainable reinforcement learning, always lean on the human capability to foresee all the consequences from simple reflexes frozen in the moment (e.g. heat maps in Atari). This can be somewhat effective in well-known everyday scenarios. Here we aim at a simple automation of task previously done by humans. We argue that even in this case, we should help people as things escalate pretty fast with the increasing complexity of the everyday world. Sudden automation that takes a locally-optimal, but counter-intuitive, decision on the grand-scale can with multiplicity (replication) create many unconceived problems. The point is driven even stronger for domains to which people are not accustomed to. It is too much to expect of humans to produce their own consequential profiles from the automated decisions. Since a consequence profile must be predictively produced for decisions, there must be some kind of anticipative/predictive model. The predictive model can be of an entangled form or a disentangled form. A disentangled model contains the model of the environment in the form of a decision stochastic process and it can model effects of different decisions, at a greater computational cost. An entangled model is in a form of a chain stochastic process where policy decisions are already entangled with the environment model. This makes them more limited, but they are computationally much cheaper. 

\noindent\textbf{Risk attitudes (\ref{enum:P2})}

The users can have different stances towards risk: avoiding, neutral, seeking. Usually, research deals with the neutral risk stance as it is computationally the simplest, especially in the case of reinforcement learning, as a normatively picked point for all the agents. Modeling anything different would necessitate a procedure that either has different risk attitudes cached in, potentially with neighborhood based interpolation between samples of attitudes, or the procedure could make adjustments for the attitudes at the input on the fly.

 \noindent\textbf{Ontological vs epistemological explanations (\ref{enum:P3})}

The problem with current methods is that there is not significant difference between ontological and epistemological perspectives. The ontological perspective addresses the case for an omniscient exact solving algorithm that has the full model of the problem and has the necessary computational resources to do calculations without making compromises. The epistemological perspective addresses a more realistic situation where an approximate-reasoning agent has neither all the information nor computational resources necessary to calculate all entailments. In the latter we must take into account not just the problem but also the algorithms that were used and the knowledge that the algorithms have about the solution. All non-exact solving procedures entangle the problem and solution procedure since approximate methods omit some considerations during the execution. These considerations must be also taken into account during the explanation. This is especially a problem from the aspect of honesty of explanation, as demonstrated by unfairness of explainability in\cite{brcic_impossibility_2022}. For example, an adversarial algorithm could make decisions based on some decision paths and scenarios that substantially shift the current-moment decisions, due to the long-term consequences. However, an explanation could potentially be constructed by forging other, benevolent  information that leads to the same current decision, based on similar foreseeable short-to-medium term consequences. Epistemological concerns make the process of honest explaining more complex.

We hold that an idealized explanation algorithm should, for simplicity, address the ontological position first – the explanation of decisions made by exact algorithms in their respective domains. The benefit of the ontological perspective is that solving method is disentangled from the problem and the two can be analyzed separately. This also lets us set the standard for all approximate methods, by potentially relaxing the explanation procedure for the ontological case and appending it with additional considerations for actual knowledge (epistemology) of an agent.

\subsection{Examples}

In this section, we shall demonstrate our proposal on simple environments. Herein we will offer the ontological explanation of a decision if the agent is capable of doing an exact optimization disregarding the resource constraints, but incorporating the insufficient knowledge and learning. Regarding the axes introduced in this paper, we set the following choices: post hoc, local, and proactive explanations. 

\noindent On the remaining three axes we shall cover all the combinations:
\begin{enumerate}[label=\textbf{A\arabic*}, leftmargin=*]
\setcounter{enumi}{3}
    \item Type of environment
       \begin{enumerate}[label=(\alph*)]
         \item deterministic
         \item stochastic
       \end{enumerate}
    \item Agent cardinality
        \begin{enumerate}[label=(\alph*)]
         \item single-agent
         \item multi-agent
       \end{enumerate}
    \item Type of policy
        \begin{enumerate}[label=(\alph*)]
         \item deterministic
         \item stochastic
       \end{enumerate}
\end{enumerate}

We propose the future research to then take into account explaining the perspective of an agent taking into account its algorithm. The explanation regarding "why" and “how” the decision has been made can significantly depend on the details beyond the explanation of a mere decision. The explanation should take into account that a decision is made “to the best of knowledge” of exploration strategies of DM algorithms.

We shall use a similarly adjustable problem for showing examples and necessary explanation elements. Explanations in examples will aim at visualizations and symbolic expressions. Let us examine the problem of crossing the shortest path between two points. With respect to adjustability according to the axes \ref{enum:A4}-\ref{enum:A6}, we will show explanations on examples:

\begin{enumerate}[label=\textbf{E\arabic*}, leftmargin=*]
    \item \label{enum:E1}
    deterministic, single-agent, deterministic – this is a well-known situation that is solved by Dijkstra’s shortest path algorithm. The explanation can be the shortest path with a contrasting list of top alternative shortest paths + mathematical proof that the given solution is the shortest path. The solution needs not even encode a complete policy, just the sequence of decisions (\ref{enum:P1}). There are no risks involved (\ref{enum:P2}). Epistemological concerns have no importance as all the information is known, and the algorithm is able to take it into account efficiently (\ref{enum:P3}).
    \item \label{enum:E2}
    Stochastic, single-agent, deterministic – such a situation is optimal in complete observability. The policy can be simulated many times and give heatmaps of rollouts for each part of the network in form of how many scenarios has been traversed. Rollout heatmaps can possibly also be done for other choices of the first decision and complete rollout with an optimal policy (a-la Q-learning). Contrasting rollout maps can address the proactivity of explanation (\ref{enum:P1}). Risks can be given in the explanation by the probability distribution of rollout lengths from the current state, especially given the probability distribution shape and its measures of risk (\ref{enum:P2}). Assuming an exact solving procedure for MDPs (in the case of complete observability), epistemological concerns are not related to the DMs algorithm, but to the uncertainty in the environment which is addressed already with elements addressing (\ref{enum:P1}) and (\ref{enum:P2}). However, in the case of a partially observable environment, an explanation procedure needs to address partial observability limitations imposed onto it by having to use a deterministic policy, which is suboptimal. An explanation procedure must warn about its limitations and what losses of performance might be involved for each decision (\ref{enum:P1}+\ref{enum:P2}). However, regarding \ref{enum:P3}, there is still the solitary limitation of constraints on external, not intrinsic constraints of agent. The limit on deterministic policies was exogenous. The used algorithm for solving within such a constraint exactly enables a disentangled explanation with the agent’s full (saturated) knowledge of the potential consequences of choices.
    \item \label{enum:E3}
    Stochastic, single-agent, stochastic – will drop down to deterministic policy for an optimal solution, if we are in the above case of complete observability. In partial observability conditions, the stochastic policy may be optimal. This is a case similar to the previous one in respect to dealing with \ref{enum:P1} and \ref{enum:P2}. However, the agent is not limited to the use of a deterministic policy and hence, does not need to warn of its limitations in that regard. The distribution over actions must be explained itself, symbolically and through visualizations. This is especially important through the explanation of expectation on small samples. That is, expectation maximizes average performance on many trials, but we often make only one trial. As the agent can calculate an optimal solution, it has no need to address \ref{enum:P3} as it obtains an ontological perspective which disentangles the solving method and the problem.
    \item \label{enum:E4}
    Deterministic, multi-agent, deterministic – where an adversary can at each step select one edge on which traversal is prolonged by a fixed amount. Optimal policy can be calculated to optimality when constrained to search among deterministic policies. As it was assumed at the beginning of the section, the agent has no limitations beyond the ones stated here. However, deterministic policy is a hard constraint on some environments. This case possibly necessitates a mixed (randomized) policy, even though all other aspects of the problem are deterministic. In the case of simultaneous decision making by players, a mixed policy might be necessary. In turn-based decision-making (such as chess) a deterministic policy might be sufficient. This is also similar to the explanations of the \ref{enum:E2} case. The difference here is that we have no indifferent environment, but biased agents (adversarial, cooperative or combinations). Hence, simulation traces with responses from an idealized opponent (or the perfect model of sub-ideal opponent) would again be a good explanation tool: rollout heatmaps with explanations of other agent’s decisions at each point, local sensitivity analysis on the current decision (coupled by other agent’s responses) (\ref{enum:P1}), sensitivity analysis for different risk attitudes (\ref{enum:P2}) and an explanation of effect of exogenous limitations on policy space. However, we again need no appeal to epistemic problems as we assume using an optimal algorithm for solving a problem described by exogenous limits. 
    \item \label{enum:E5}
	Deterministic, multi-agent, stochastic – is similar to the above scenario. In this case, the limit on the solution space is reduced so the aspect of that exogenous limitation needs not be addressed. However, the decisions of stochastic policy must be explained in the regime of small samples (\ref{enum:P1}+\ref{enum:P2}).
    \item \label{enum:E6}
    Stochastic, multi-agent, stochastic – here an adversary makes choices of prolonging some edges, which the environment might fill with some noise. Also, the traversal of the edge might be uncertain itself (even without the added prolongation). This is the most complex situation for giving explanations as there is so much unknown in the problem, and it builds upon the previous situation. There is uncertainty in the agent’s own choices, environment’s reaction, and other agent’s decisions. Again, the visualization and symbolic expressions based on simulation traces with idealized opponent (or perfect model of sub-ideal opponent) are the main proposed tool (\ref{enum:P1}). We need to address both components of exogenous uncertainty. Environment reactions need no explanation, but other agent’s do, to a degree. Sensitivity analyses over current decision (\ref{enum:P1}) and risk attitudes (\ref{enum:P2}) deal with additional aspects. 
    
\end{enumerate}

We should be aware of the possibility that multi-agent problems might include situations where consistency is fleeting and such a case needs to be addressed in explanations as an appeal to mathematical impossibilities \cite{brcic_impossibility_2022}. What might it mean for solving and explaining the solutions where an ontological perspective does not even exist? Furthermore, we can see that simulation traces necessitate some process of producing them, and the open question is if that can be circumvented. Reactive explanations of choosing a current decision in a graph would not account for seeing full consequences of future actions, since they aggregate the whole future into a current momentary decision. Not accounting for risk fails to adjust decision-making to different decision-makers’ profiles. Not accounting for the ontologic-epistemic gap in approximate algorithms fails to address the risks due to the way solving procedures work in a way of:
\begin{itemize}
    \item missed opportunities,
    \item unforeseen risks,
    \item hidden adversarial behavior that is hidden behind faked epistemic limits.
\end{itemize}

\section{Conclusion}
\label{sec:concl}
In this paper, we have shown the current state of the art in the explainable RL area. In addition to the classical categorization according to the timing and scope of explanation, we added four new axes for categorization and description of existing approaches: time horizon of explanation (reactive/proactive), type of environment (deterministic/stochastic), type of policy (deterministic/stochastic) and agent cardinality. With our first new axis, we want to emphasize the importance of a proactive explanation. It is shown that human users tend to favor explanations about policy rather than about a single action \cite{van_der_waa_contrastive_2018}. 
We have listed current work categorized in several groups: Structural causal model, explanation in terms of consequences, hierarchical policy, relational reinforcement learning, policy simplification, reward decomposition. For each group, we gave a description and listed its advantages and disadvantages (see Table \ref{tab:groups_subgroups}).

One of the long standing challenges of utilizing XAI techniques in general are the difficulty to evaluate (i) is a provided explanation correct or not \cite{chen_seven_2021} and (ii) what are the unanticipated negative downstream effects of a given explanation or possible explainability pitfalls in general\cite{ehsan_explainability_2021}. Therefore, the adaptation of XAI techniques (originally designed for supervised settings) to XRL approaches, make the challenges to inherently transition into the XLR domain, which demands for novel conceptual framework for assessment of XRL methods in various application scenarios with respect to target users and their domain problems \cite{sokol_explainability_2020}. Our proposal also sets a foundation for development of this conceptual framework in form similar to Explainability Fact Sheets introduced in  \cite{sokol_explainability_2020}.

People want to understand the policy that is the baseline to the decision-making process.
Most methods only offer an explanation of the current decision, and the whole decision-making process remains unexplained. When solving large problems, the number of explanations becomes inordinate. Due to limited computer resources and limited human memory, a compromise is made at the expense of accuracy. This problem is also solved by limiting the scope of the problem (predefined number of states and actions), but makes the model inapplicable to the whole area. Also, most explanations are intended for experts in a field, or at least human engagement in generating explanations is required.

Finally, we aim to set standards for explanation in order to promote more ethical behavior and wide-scale cooperation. Our position for truthful explanations has three pillars: explanations should be proactive, explanations must take into account risk attitudes and decision-maker’s computational process must be taken into account in the form of epistemological constraints. We illustrated our proposal on simple variants of the shortest path problem. Each variant was created through variation of our proposed axes: type of environment, agent cardinality and policy type. For each variant we have given a sketch of how to address each of our three pillars. 

Future work should address ontological perspective more precisely since it decouples an agent’s algorithm from the problem. Only then we should deal with greater and general forms of entanglement between the two. In more realistic use-cases some information might be available in principle, but it is practically unobtainable. For example, this may be the case due to the constraints on computational resources of time and space when tradeoffs must be done in approximation procedures. In those cases, we must investigate the sensitivity of validity of explanation with regards to the information omitted (from a knowledge of agent or from an explanation form).

\bibliographystyle{IEEEtran}
\bibliography{IEEEabrv,XRL}

\end{document}